\documentclass[conference]{IEEEtran}
\IEEEoverridecommandlockouts
\usepackage{amsmath,amsfonts,amssymb,amsthm}
\usepackage{verbatim}
\usepackage{color}
\usepackage{epsfig}
\usepackage{array}
\usepackage{algorithm}
\usepackage{cite}
\usepackage{tabu}
\usepackage{mathtools}

\usepackage{wrapfig}
\usepackage{MyMnSymbol}
\usepackage{algorithm}
\usepackage[noend]{algpseudocode}
\usepackage{url}
\usepackage{placeins}
\usepackage[super]{nth}
\def\BibTeX{{\rm B\kern-.05em{\sc i\kern-.025em b}\kern-.08em
    T\kern-.1667em\lower.7ex\hbox{E}\kern-.125emX}}

\begin{document}

\title{Volumetric Parameterization of the Placenta to a Flattened Template}

\author{S. Mazdak Abulnaga, Esra Abaci Turk, Mikhail Bessmeltsev, P. Ellen Grant, Justin Solomon, and Polina Golland
\thanks{This work was supported in part by NIH R01HD100009, NIH U01HD087211, NIH R01EB017337, NIH NIBIB NAC
P41EB015902, Wistron Corporation, ARO W911NF2010168, AFOSR FA9550-19-1-031, NSF IIS-1838071, NSF CHS-1955697, NSERC RGPIN-2019-05097, FRQNT 2020-NC-270087, NSF GRFP, NSERC PGS D, and MathWorks Fellowship.}
\thanks{S. M. Abulnaga, J. Solomon, and P. Golland are with the Computer Science and Artificial Intelligence Lab, Massachusetts Institute of Technology, Cambridge, 02139, USA (e-mail: abulnaga@mit.edu).}
\thanks{E. Abaci Turk and P. E. Grant are with the Fetal-Neonatal Neuroimaging and Developmental Science Center, Boston Children's Hospital, Harvard Medical School, Boston, MA, 02115, USA.}
\thanks{M. Bessmeltsev is with the Department of Computer Science and Operations Research, Universit\'{e} de Montr\'{e}al, Montr\'{e}al, QC, H3C 3J7, Canada.  }}

\maketitle

\begin{abstract}
We present a volumetric mesh-based algorithm for parameterizing the placenta to a flattened template to enable effective visualization of local anatomy and function. MRI shows potential as a research tool as it provides signals directly related to placental function. However, due to the curved and highly variable \textit{in vivo} shape of the placenta, interpreting and visualizing these images is difficult. We address  interpretation challenges by mapping the placenta so that it resembles the familiar \textit{ex vivo} shape. We formulate the parameterization as an optimization problem for mapping the placental shape represented by a volumetric mesh to a flattened template. We employ the symmetric Dirichlet energy to control local distortion throughout the volume. Local injectivity in the mapping is enforced by a constrained line search during the gradient descent optimization. We validate our method using a research study of 111 placental shapes extracted from BOLD MRI images. Our mapping achieves sub-voxel accuracy in matching the template while maintaining low distortion throughout the volume. We demonstrate how the resulting flattening of the placenta improves visualization of anatomy and function. Our code is freely available at \url{https://github.com/mabulnaga/placenta-flattening}.

\begin{IEEEkeywords}
Anatomy visualization, injective map, fetal MRI, placenta, shape, volumetric mesh parameterization.
\end{IEEEkeywords}

\end{abstract}
\section{Introduction}
\IEEEPARstart{W}{e} present a volumetric method for mapping the placenta to a flattened template to enable examination of local placental anatomy and function. The placenta is an organ that connects the fetus to the maternal blood system to provide oxygen and nutrients to enable fetal growth~\cite{benirschke1967pathology}. Placental insufficiency can affect the mother's health and cause long-term problems for a child's health and development~\cite{benirschke1967pathology,turk2019placental}. 
Blood oxygen level dependent (BOLD) MRI has been used as a research tool to quantify oxygen transport within the placenta~\cite{sorensen2013changes,luo2017vivo,pratt2015computer,turk2019placental,schabel2016functional,abaciturk2020placenta}, providing initial 
evidence for clinical utility of MRI for functional assessment. Further, T2-weighted MRI captures anatomical information such as the vascular structure of the placenta~\cite{torrents2019fully}. Since MRI provides direct measurements of placental function, it is a promising research tool for studying the placenta~\cite{sorensen2013changes,luo2017vivo,turk2019placental,
abaciturk2020placenta,sorenson2020t2,ho2020t2}.

MRI has been used to study placental oxygen transport~\cite{turk2019placental,hutter2018function}, fetal-maternal circulation~\cite{melbourne2019circulations}, uterine contractions~\cite{abaciturk2020placenta}, and to characterize the relationship between placental age and its appearance in MRI~\cite{pietsch2021applause}. A promising research direction is to use MRI  to develop biomarkers of pathology. BOLD and T2* MRI show promise for identifying intrauterine growth restriction (IUGR)~\cite{sinding2016placental}, for quantifying differences in discordant twins with IUGR~\cite{luo2017vivo}, and for characterizing preeclampsia~\cite{ho2020t2}. While emerging as a promising tool to study placental function and health, effective visualization of MRI signals in the placenta is an open problem.

Interpreting MRI scans of the placenta is challenging due to its \textit{in vivo} shape. The placental shape is determined by the curved shape of the uterine wall, making it difficult to visualize the interior of the organ using standard techniques, for example, using cutting planes. As the location of the placenta's attachment to the uterine wall, its shape, and its appearance vary greatly across subjects, a standard visualization of the functional and anatomical images of the placenta is yet to be demonstrated.

As part of the post-delivery examination process, the placenta is placed on a flat surface, and measurements of its disk-like shape and notes about its anatomical features are recorded~\cite{gersell1998ascp}. Motivated by this framework for examination of the placenta, we present an algorithm for mapping the placental shape observed in an MRI scan to a flattened template that resembles the organ's well-studied \textit{ex vivo} shape.

We represent the placental volume as a tetrahedral mesh and find the deformation to a flattened template as the solution to an optimization problem. We provide a general formulation that readily accepts a broad set of templates and shape distortion measures. We compute an injective map by constraining the minimization of the objective function, to ensure that anatomical fidelity is preserved. In this paper, we select two parallel planes as our flattened template surface and measure shape distortion by the symmetric Dirichlet energy to penalize local deformations~\cite{schreiner2004inter}. We propose and evaluate alternative templates including an ellipsoid and a relaxation of the parallel planes template. Moreover, we provide guidelines for template selection based on the desired visualization or computation task. 

We evaluate the proposed method by applying it to segmented BOLD images of a research study. We demonstrate effective mapping of the highly variable placental shapes in our dataset. We highlight the utility of our method for research by presenting enhanced visualization of local anatomical and functional features, and demonstrate the possibility of quantifying MRI signals within functional regions of the placenta.

\subsection{Related Work}
We build on state-of-the-art mesh parameterization methods to estimate the optimal deformation of the placenta onto a template. Parameterization is the problem of mapping from a shape to a parameter domain, for example, from a surface mesh to a plane~\cite{timsari2000optimization,smith2015bijective,floater2003mean,
tutte1963draw,rabinovich2016scalable,hormann2000mips,
fu2015amips,mullen2008spectral,levy2002least,
desbrun1999implicit,miao2017placenta,halier2000nondistorting,
haker2000colon,desbrun2002intrinsic} 
 or to a sphere~\cite{gotsman2003fundamentals,haker2000conformal,
lui2007landmark,fischl1999cortical,fischl2012freesurfer,tosun2001hemispherical}. In geometry processing, parameterization methods are useful for applications such as texture mapping, smoothing, and remeshing~\cite{sheffer2007mesh}. In medical imaging, parameterization has been used to improve visualization of anatomy and to facilitate population studies~\cite{kreiser2018survey}. Surface-based parameterization has been used in many applications including cortical surface mapping~\cite{haker2000conformal, lui2007landmark, 
  fischl1999cortical,timsari2000optimization,tosun2001hemispherical,fischl2012freesurfer},
  heart and vessel flattening~\cite{oeltze2006integrated, termeer2007covicad}, colon unfolding~\cite{halier2000nondistorting,haker2000colon}, bone unfolding~\cite{bruckner2017bone,vrtovec2007automated}, and placental flattening~\cite{miao2017placenta}. For a more detailed treatment of parameterization methods in geometry processing and in medical imaging, we refer the reader to surveys~\cite{hormann2008mesh,floater2005surface,sheffer2007mesh,kreiser2018survey}. In contrast to most algorithms that focus on mapping 2D surface meshes, we address the parameterizion of volumetric meshes to a 3D canonical template volume.
  
While surface parameterization methods are not directly applicable to volumetric meshes, they provide a framework for computing the desired bijective map with minimal distortion. Surface parameterization approaches use Tutte's embedding to guarantee a bijective planar parameterization through a linear map to a closed, convex boundary~\cite{tutte1963draw,floater2003one}.  A bijective spherical parameterization can be achieved using Tutte's embedding with an inverse stereographic projection~\cite{haker2000conformal,tosun2001hemispherical} or by using one of several proposed extensions of the embedding to the sphere~\cite{gotsman2003fundamentals,aigerman2017spherical}. As this initial parameterization is likely to have high distortion, the map is refined by minimizing a cost function that includes a distortion metric to achieve a parameterization with desirable properties such as bijectivity and minimal distortion. 
Existing methods optimize distortion metrics to compute an angle-preserving (conformal) map~\cite{mullen2008spectral,levy2002least,haker2000conformal,lui2007landmark,
tosun2001hemispherical,halier2000nondistorting,haker2000colon,vrtovec2007automated}, an area-preserving (authalic) map~\cite{desbrun2002intrinsic,timsari2000optimization}, a distance-preserving (isometric) map~\cite{rabinovich2016scalable,hormann2000mips,fu2015amips,
fischl1999cortical,timsari2000optimization}, or a map that minimizes a combination of undesired distortions~\cite{rabinovich2016scalable,desbrun1999implicit,schreiner2004inter,ezuz2019reversible}. To achieve a bijective map, a barrier functional is used to avoid flipped triangles and boundary intersections~\cite{smith2015bijective}.

In the volumetric case, the existence of an exact bijective parameterization is not guaranteed~\cite{aigerman2013injective,floater2006convex}. The challenge is to enforce injectivity when mapping to a defined parameterized domain. Many of the distortion measures defined in 2D are easily extendable to 3D. Aigerman \textit{et al}.~\cite{aigerman2013injective} develop an algorithm that projects an existing map to the space of injective and bounded-distortion maps. They demonstrate successful parameterization to a polycube domain. In contrast, we enforce injectivity at every step of the optimization to explicitly control distortion while maintaining an injective map. We use gradient descent with line search to prevent tetrahedra from folding. This line search method was originally proposed for 2D parameterization~\cite{smith2015bijective} and has been effective for 3D deformation of tetrahedral meshes of cubes~\cite{rabinovich2016scalable,fu2015amips} and animal shapes in computer graphics~\cite{fu2015amips}. 
Here, we minimize local distortion using the 3D extension of the symmetric Dirichlet energy~\cite{schreiner2004inter}. We propose a problem-specific unique volumetric template to serve as our parameterized domain during the optimization. We demonstrate successful parameterization of a dataset of highly variable placental shapes. 

Most closely related to our work is the method by Miao \textit{et al}.~\cite{miao2017placenta} that represents the placenta volume as a stack of parallel surfaces spanning the thickness of the organ, spaced at discrete intervals. The surfaces are estimated based on the medial axis of the placenta. Each surface is flattened independently using mean value coordinates~\cite{floater2003mean} to iteratively move each interior vertex to the average of its neighbors and to map the surface to a disk. There are two limitations to this approach. First, due to the highly variable shape of the placenta, parameterizing to a fixed boundary leads to high distortion, affecting important image features. Second, because the 2D surfaces are each parameterized independently, no correspondence or alignment across layers is enforced, which distorts important depth-wise image features. In contrast, we propose a continuous volumetric parameterization without a fixed boundary, ensuring consistency and minimal distortion throughout the placental volume. Our work presents a standardized method for visualizing signals within the placenta, and is the first step towards developing a common coordinate system for placental analysis.

A preliminary version of our method was presented at the International Conference on Medical Image Computing and Computer-Assisted Intervention (MICCAI) 2019~\cite{abulnaga2019placenta}. Here, we derive the parameterization in greater detail by expanding on the definition of our template and the optimization procedure. We provide a template formulation that is robust to irregularities in the meshing procedure and extensions to closely match the \textit{ex vivo} shape. We demonstrate our algorithm on a significantly larger research dataset of 111 placental shapes. Finally, we conduct extensive experiments and demonstrate that the map is robust to placental shape, produces uniform distortion, and improves visualization of local anatomy and function. 

This paper is organized as follows. In the next section, we formulate the optimization problem for parameterizing placental volumes to our proposed flattened template, outline the optimization procedure to compute an injective map, and provide implementation details. In Section~\ref{s:experiments}, we introduce the BOLD MRI data and report the experimental results and comparisons with the baseline method. In Section~\ref{s:discussion}, we discuss potential applications of our method, outline limitations in using the proposed template as a common space, and present future directions to incorporate anatomical information to create a common coordinate system. Our conclusions follow in Section~\ref{s:conclusion}.

\section{Methods}

We model the placental shape as a tetrahedral mesh that contains $N$~vertices and $K$~tetrahedra. In our experiments, we extract the mesh from the placenta segmentation in an MRI scan as described in Section~\ref{s:data}. We parameterize the mapping via mesh vertex locations and interpolate the deformation to the interior of each tetrahedron using a locally affine (piecewise-linear) model. As the mapping is geometry-based, it is independent of the imaging modality used to obtain the input mesh.

\begin{figure}[!t]
\centering
\includegraphics[width=0.49\textwidth]{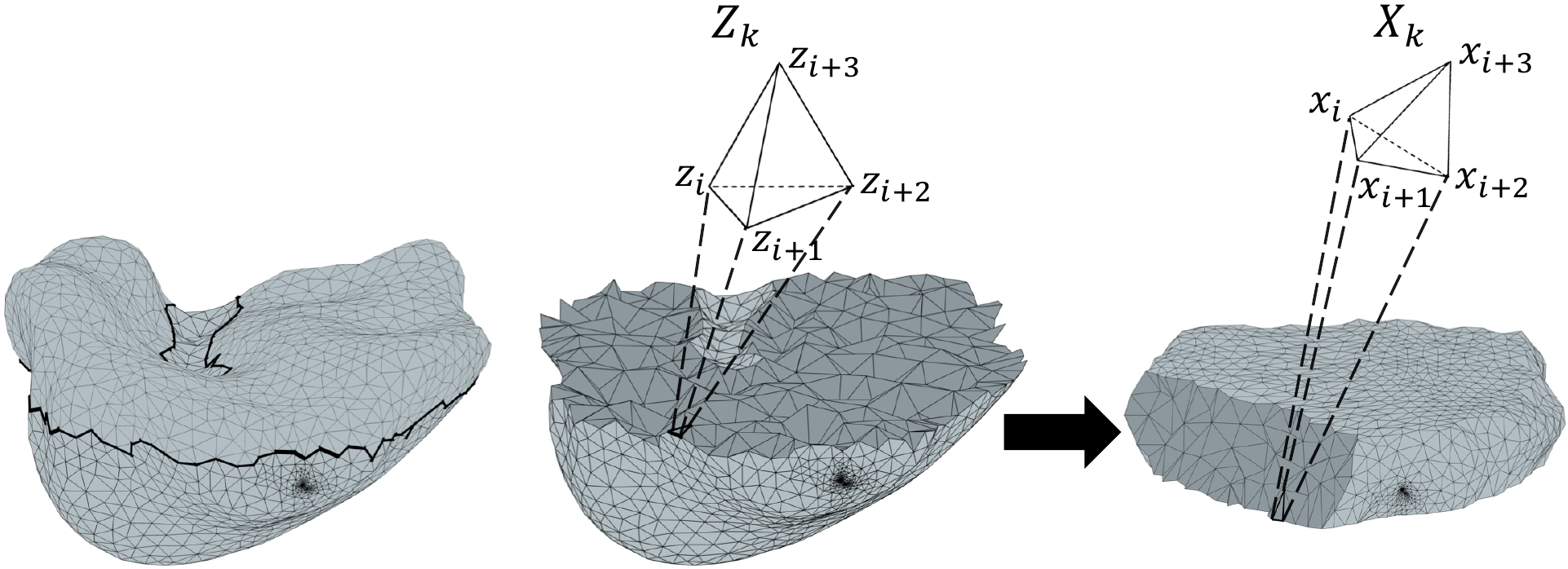}
\caption{Visual depiction of the mapping procedure. Left: original placental mesh, with boundary triangles shown. The solid black line indicates an area that will be cut to illustrate the interior tetrahedra. Center: the original placental mesh after cutting the solid lined-area, with the interior tetrahedra shown in dark gray; right: the resulting flattened placental mesh, depicting the interior tetrahedra. We use \textit{Z} and \textit{X} to denote the variables in the original image space and the template space respectively.}
\label{f:mesh-diagram}
\end{figure}

\subsection{Problem Formulation}
Let $x_i \in \mathbb{R}^{3}$ define the 3D coordinates of mesh vertex $i$ in the template coordinate system, and let $X = \{x_1 \dots x_N\} \in \mathbb{R}^{3\times N}$ be the coordinates of all vertices in the mesh. Tetrahedron~$k$ is represented by a matrix $X_k \in \mathbb{R}^{3\times 4}$ whose columns are the 3D coordinates of the four corner vertices of tetrahedron $k \in \{1,\ldots,K\}$ in the template space. Similarly, we use $Z \in \mathbb{R}^{3\times N}$ to denote the coordinates of mesh vertices $\{z_i\}_{i=1}^{N}$ in the original image space, and $Z_k \in \mathbb{R}^{3\times 4}$ to define tetrahedron~$k$ in the original image space. We use $\partial$ to denote the boundary of a mesh, where $\partial Z$ defines the set of vertex indices of $Z$ that lie on the mesh boundary. The elements of $\partial Z$ remain constant throughout the parameterization. Fig.~\ref{f:mesh-diagram} provides a visualization of the flattening procedure and the notation. 

We formulate placental parameterization as an optimization problem over the mesh vertices that seeks to map the vertices to the template space while minimizing a measure of local distortion: 

\begin{equation}
\label{eqn:objectivefun}
\phi(X,\theta)= 
\underbrace{\sum_{m \in \partial Z } A_m \: T\left(x_m, \theta \right)}_{\text{Template match}}  + \underbrace{\lambda \sum_{k=1}^K V_k \:
\mathcal{D}\left(X_k\right)}_{\text{Distortion}},
\end{equation}

\noindent where $x_m$ denotes the coordinates of boundary vertex $m$, function $T\left(\cdot\right)$ measures the distance of the boundary vertices to the template shape, $\theta$ represents the parameters defining the template, $A_m$ is a constant defined to be the normalized barycentric area associated with vertex $m$ on the boundary of the mesh (i.e., $A_m$ is proportional to the sum of
the areas of the triangle faces that include vertex $m$ and $\sum_{m \in \partial Z} A_m=
1$), $V_k$ is a constant defined to be the normalized volume of tetrahedron $k$ in the original shape space ($\sum_{k=1}^K V_k =
1$), and $\lambda$ is a parameter that governs the trade-off between the template match and the shape distortion. We define the specific terms of the cost function below.

\subsection{Shape Distortion}
The formulation of the distortion term in Eq.~\eqref{eqn:objectivefun} naturally accepts a range of distortion metrics provided they are differentiable in vertex positions $X$. The distortion term regularizes the mapping. 

We model the deformation of tetrahedron $Z_k$ in the original image space to its image $X_k$ in the template space as an affine transformation. The Jacobian matrix

\begin{equation}
\label{eqn:jacobian}
  J(X_k) = \left(X_k B \right) \left(Z_k B \right)^{-1}
\end{equation}
defines the deformation of tetrahedron $k$ based on the new vertex coordinates $X_k$ while ignoring the shared translation component.
The constant matrix $B\in \mathbb{R}^{4 \times 3}$ extracts three basis
vectors defining the tetrahedron; the vectors originate from the first
corner vertex of the tetrahedron and point to its other three corner
vertices (see Appendix~\ref{appendix:jacobian} for the definition of $B$).

We measure tetrahedron-specific distortion using the symmetric Dirichlet energy density
\begin{align}
\label{eqn:dirichletenergy}
\mathcal{D}(X_k) &= \mathcal{D}\left(J\left(X_k \right)\right) \nonumber \\
& =\| J \left(X_k\right)\|_{F}^{2} + \|\left(J\left(X_k\right)\right)^{-1}\|_{F}^{2} = \sum_{i=1}^{3}\left(\sigma_{k,i}^2 + \sigma_{k,i}^{-2}\right)
\end{align}
where $\| \cdot \|_{F}$ is the Frobenius norm and $\{\sigma_{k,1},
\sigma_{k,2}, \sigma_{k,3}\}$ are the singular values of matrix $J(X_k)$~\cite{rabinovich2016scalable,smith2015bijective,schreiner2004inter}.
We chose the symmetric Dirichlet energy as it penalizes expansion and shrinking equally
and prevents tetrahedra from expanding or shrinking
dramatically as the corresponding singular values or their reciprocals become unbounded. This distortion energy favors a locally injective map by preventing tetrahedra from collapsing to a point.

\begin{figure*}[h!]
\centering
\includegraphics[width=0.9\textwidth]{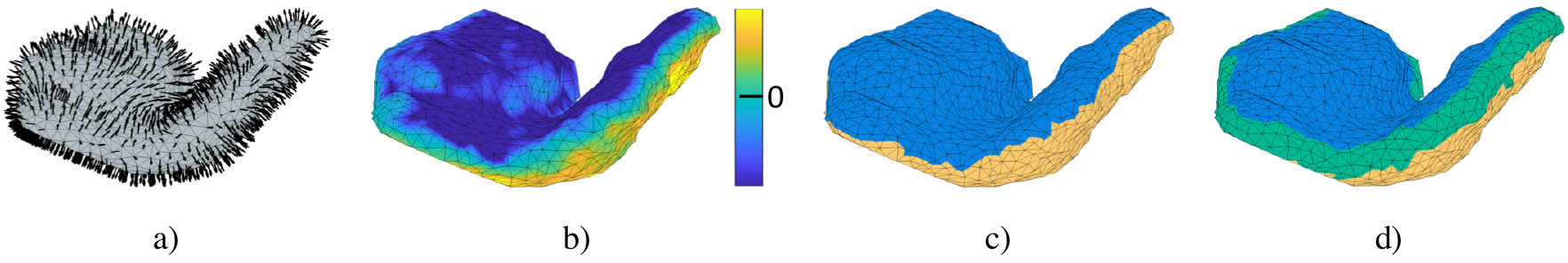}
\caption{Placenta parcellation procedure: a) placenta mesh with boundary vertex normals shown; b) spectral embedding of the second smallest eigenvector of the Laplacian $L$; c) placenta boundary segmented into the maternal (yellow) and fetal (blue) sides. The placenta margin is the set of vertices on the edge of the two clusters; d) the expanded placenta margin is shown in green. }
\label{f:spectral-cluster}
\end{figure*}

\subsection{Template Definition}
We propose two template parameterizations to resemble the \textit{ex vivo} placental shape. The first contains two parallel planes to enforce a uniform thickness, and the second models the placental shape as an ellipsoid.
\subsubsection{Parallel Planes Template}
We construct the \textit{parallel planes} template to mimic the post-delivery examination of the placenta where the maternal side of the organ is placed flat on an examination table. We define the template as two parallel planes, one for the maternal side and and one for the fetal side of the organ. The two planes ensure a constant thickness of the placenta, which facilitates 2D visualization as we demonstrate later in the paper. Function $T(\cdot)$ measures the distance to the appropriate plane:

\begin{equation}
\label{eqn-template-term}
T(x_m,h) =
\begin{cases}
(x_m^{(3)} - h)^2 & \text{if } m \in \mathcal{F}(\partial Z), \\
(x_m^{(3)} + h)^2 & \text{if } m \in \mathcal{M}(\partial Z), \\
0 & \text{otherwise},
\end{cases}
\end{equation}
where $h$ is the template half-height ($\theta = h$ in Eq.~\eqref{eqn:objectivefun}), $x_m^{(3)}$ is the third coordinate of vertex $m$ in the template coordinate system, and $\mathcal{F}(\partial Z)$, $\mathcal{M}(\partial Z) $ denote the subset of vertex indices of the fetal and maternal sides of the placental boundary surface $\partial Z$, respectively. The template half-height $h$ can be treated as either a fixed parameter or a variable. We use the term ``placenta margin" to denote the region of high curvature that separates the fetal and maternal sides of the placenta. The placenta margin is not mapped to either plane; the margin vertices are driven only by the distortion term.

The template term in Eq.~\eqref{eqn-template-term} requires identifying the maternal and fetal sides of the placenta boundary. As we expect the boundary normals to cluster into two distinct groups that correspond to the two sides of the placenta, we construct a similarity measure based on the angle between boundary vertex normals and project the vertices to a 1D space using spectral embedding~\cite{ng2002spectral,von2007tutorial}, described below.

For boundary vertex $i$, we determine the unit normal  $\hat{n}_i$ as a weighted average of normals over the mesh boundary triangles that contain vertex $i$, normalized to have unit length. The weighting is proportional to the triangle areas. We construct an affinity matrix $W \in \mathbb{R}^{|\partial Z|\times |\partial Z|}_{\geq 0}$ so that

\begin{equation}
w_{i,j} = \exp\left\{\gamma\left \langle\hat{n}_i,\hat{n}_j\right \rangle l\left(x_i,x_j\right)\right\}
\end{equation}
 for any two boundary vertices $i$ and $j$ that are connected by any path containing three or fewer edges (the three-ring neighborhood), and set $w_{i,j}=0$ otherwise. The parameter $\gamma$ penalizes the local effect of variation in the orientation of the normals, $\langle 
 \cdot , \cdot \rangle$ is a dot product, and function $l(x_i, x_j)$ measures the approximate geodesic distance between vertex positions $x_i, x_j$, normalized by the maximum distance across all boundary vertices connected within a three-ring neighborhood. Smaller values of parameter $\gamma$ cause sensitivity to fast changes in the orientation of normal vectors, and would likely cluster a small region of the mesh with high variation of normals, such as a corner-like region. We approximate geodesic distance using Dijkstra's shortest path algorithm~\cite{dijkstra1959note} along boundary mesh edges, with edge weights given by the Euclidean distance. The weighting by geodesic distance corrects for meshing irregularities along the boundary. This prevents clusters forming in small but dense areas within the fetal or maternal sides of the boundary and ensures the appropriate clusters form to capture the large regions of interest.

The normalized Laplacian of $W$ is defined as 
\begin{equation}
L = I - D^{-\frac{1}{2}}WD^{-\frac{1}{2}},
\end{equation}
where $D$ is a diagonal matrix with $d_{i,j} = \sum_j w_{i,j}$ and $I$ is the identity matrix. The second smallest eigenvector of $L$ defines a spectral embedding of the boundary vertices that can be thresholded to segment the boundary into two components with consistent orientation of the normals~\cite{ng2002spectral}. After segmenting the placenta boundary, we identify the fetal and maternal sides. We use the fact that the maternal side of the placenta boundary has a more convex shape since it is attached to the curved uterine wall. We construct the convex hull of the boundary to assign the maternal label to the cluster with the larger number of vertices on the hull.

We define the margin of the placenta as the connected region separating the fetal and maternal sides. We identify vertices on the edge of the two clusters, i.e.,\ those with neighbors in the other cluster, and assign them to the margin. We use the approximated geodesic distance to define this neighborhood, and assign all vertices within a specified distance. This distance, or the margin width, is independent of the template half-height \textit{h}. Using the geodesic distance accounts for irregularities in the mesh and ensures the margin separates the two sides of the placenta with consistent width. Fig.~\ref{f:spectral-cluster} provides a visualization of the placenta parcellation procedure. The vertex labeling as belonging to the fetal side, maternal side, or the placenta margin is defined at initialization on the original shape and remains fixed throughout the optimization.

\subsubsection{Ellipsoid Template}
An ellipsoid template is proposed as a simple parameterization to model the unfolded placenta. Function $T(\cdot)$ measures the distance to an ellipsoid,
\begin{equation}
\label{eqn:template-ellipsoid}
T(x_m,R) = (x_m^TRx_m-1)^2,
\end{equation}
where $R$ are the ellipsoidal parameters, $R=\text{diag}(r_x^{-2},r_y^{-2},r_z^{-2})$ ($\theta=R$ in Eq.~\eqref{eqn:objectivefun}).

\subsection{Optimization}
\label{s:optimization}
We minimize the objective function $\phi(\cdot)$ in Eq.~\eqref{eqn:objectivefun} using gradient descent over the vertex locations and the template half-height. We initialize the mapping by the identity transformation by setting $X=Z$. The gradient of $\phi(\cdot)$ is given in Appendix~\ref{appendix:gradient}. 

We optimize using line search to compute an injective map by preventing tetrahedra from ``flipping" at every iteration~\cite{smith2015bijective}. A tetrahedron ``flips" after crossing the singularity point of zero volume, causing tetrahedra to fold onto one another. At every iteration, we determine the largest step size value $\eta$ such that moving the current vertex locations~$X$ by $-\eta \nabla \phi(X)$ avoids singularities for all tetrahedra~\cite{smith2015bijective}. The (signed) volume $V_k$ of
tetrahedron~$k$

\begin{equation}	
\label{eqn-tetvolume}
V_k \left( \eta \right) = \frac{1}{6}\det \Big[ \big ( X_k-\eta \nabla \phi ( X_k) \big ) B \Big ]
\end{equation}

\noindent is a cubic polynomial of $\eta$ whose real and positive roots are the step sizes that lead to the tetrahedron flipping. The upper limit $\eta_{max}$ in a line search is found by finding the smallest positive real root of Eq.~\eqref{eqn-tetvolume} over all tetrahedra~\cite{smith2015bijective}. We optimize using backtracking line search. At every iteration, $\eta$ is set to a fraction $\beta \in (0,1)$ below $\eta_{max}$, $\eta = \beta \cdot \eta_{max}$, and is decreased by a factor $\rho \in (0,1)$ until the new set of vertex locations decreases the objective function~\cite{nocedal2006numerical}. Since the original value of $\eta_{max}$ can result in an increase of the objective function, precluding convergence, backtracking line search is needed to ensure convergence by guaranteeing that the objective function decreases in every iteration. Algorithm~\ref{alg:mapping} describes the optimization procedure.

\begin{algorithm}
\caption{Volumetric Map with Local Injectivity}\label{alg:mapping}
\begin{flushleft}
        \textbf{Input:} $Z,\theta,\lambda$\\
        \textbf{Output:} $X$ 
\end{flushleft}
\begin{algorithmic}[1]
\State $X^{(0)} \gets Z$
\State $\theta^{(0)} \gets {\theta}$
\State $\phi^{(0)} \gets  \phi(X^{(0)},\theta^{(0)}, \lambda$) \textit{// initial objective function}

\State $\nabla_X \phi^{(0)} \gets \nabla_X \phi(X^{(0)}, \theta^{(0)}, \lambda)$ \textit{// initial gradient}
\State $l \gets 1$ \textit{// iteration counter} 

\While{ $\|\nabla_X \phi^{(l-1)}\|_F > \epsilon$}
\State $\eta_{max} \gets \min\left( \text{Eq.} \eqref{eqn-tetvolume}\right)$ \textit{// over all tetrahedra}
\State $\eta \gets \beta \cdot \eta_{max}$

\State $\textit{X}^{(l)} \gets X^{(l-1)} - \eta \nabla_X \phi^{(l-1)}$
\State $\theta^{(l)} \gets \theta^{(l-1)} - \eta  \nabla_{\theta} \phi^{(l-1)}$
\State $\phi^{(l)} \gets  \phi(X^{(l)},\theta^{(l)})$
	\While{$\phi^{(l)} \geq \phi^{(l-1)}}$ \textit{// backtracking line search}
		\State$\eta \gets \eta \cdot \rho$
		\State $\textit{X}^{(l)} \gets X^{(l-1)} - \eta \nabla_X \phi^{(l-1)}$
		\State $\phi^{(l)} \gets  {\phi(X^{(l)},\theta^{(l)}})$
	\EndWhile
\State $l \gets l+1$
\EndWhile
\State $X \gets X^{(l)}$
\end{algorithmic}
\end{algorithm}

\subsection{Intensity Mapping}
\label{s:intensity-map}
Once the optimal vertex coordinates are found, we map
image intensity inside the placenta segmentation to the template
coordinate system. Since our mapping is injective and piecewise affine, we use barycentric coordinates to determine the
unique transformation for any point inside the placenta. Specifically, point $x$ inside tetrahedron $k$ can be represented as a convex
combination of the tetrahedron's corner vertices, \textit{i.e.,} $x = X_k \alpha$, through the point's
barycentric coordinates $\alpha$ as follows:
\begin{equation}
  \left[ \alpha_2 \;\; \alpha_3 \;\; \alpha_4\right]^T  = \left(X_k B\right)^{-T} \left( x - X_k^1 \right),  
\end{equation}
where $X_k^1$ is the first column of matrix $X_k$. Setting $\alpha_1 =
1 - (\alpha_2 + \alpha_3 + \alpha_4)$ completes the vector to ensure $\sum_{i=1}^4 \alpha_i = 1$ \cite{mobius1827bary}. The point's coordinates $z = Z_k\alpha$ in the original volume are used to pull the image intensity into the template
coordinate system using linear interpolation.

\subsection{Implementation Details}

We generate tetrahedral meshes from segmentation labelmaps using
iso2mesh~\cite{fang2009tetrahedral}, a MATLAB toolbox that
provides an interface to TetGen~\cite{si2006quality}, a commonly used
tetrahedral meshing library. Prior to applying the mapping, we center
the mesh and rotate it to align its principal axes to the
axes of the template. We assume that the algorithm convergences when the Frobenius norm of the gradient is lower than $\epsilon = 10^{-4}$.

We implemented the algorithm in MATLAB using GPU functionality to
parallelize the computation of the gradient and the line search. We
ran our experiments on an NVIDIA Titan Xp (12GB) GPU. The parameterization ran for an average of 5097 iterations for the parallel planes template and 7203 iterations for the ellipsoid template on our dataset of meshes containing a mean of 6708 tetrahedra. The mean and \nth{90} percentile wall clock times were $17$ and $31$ minutes respectively for the parallel planes template and $26$ and $49$ minutes respectively for the ellipsoid template.

\begin{figure}[!t]
\centering
\includegraphics[width=0.4\textwidth]{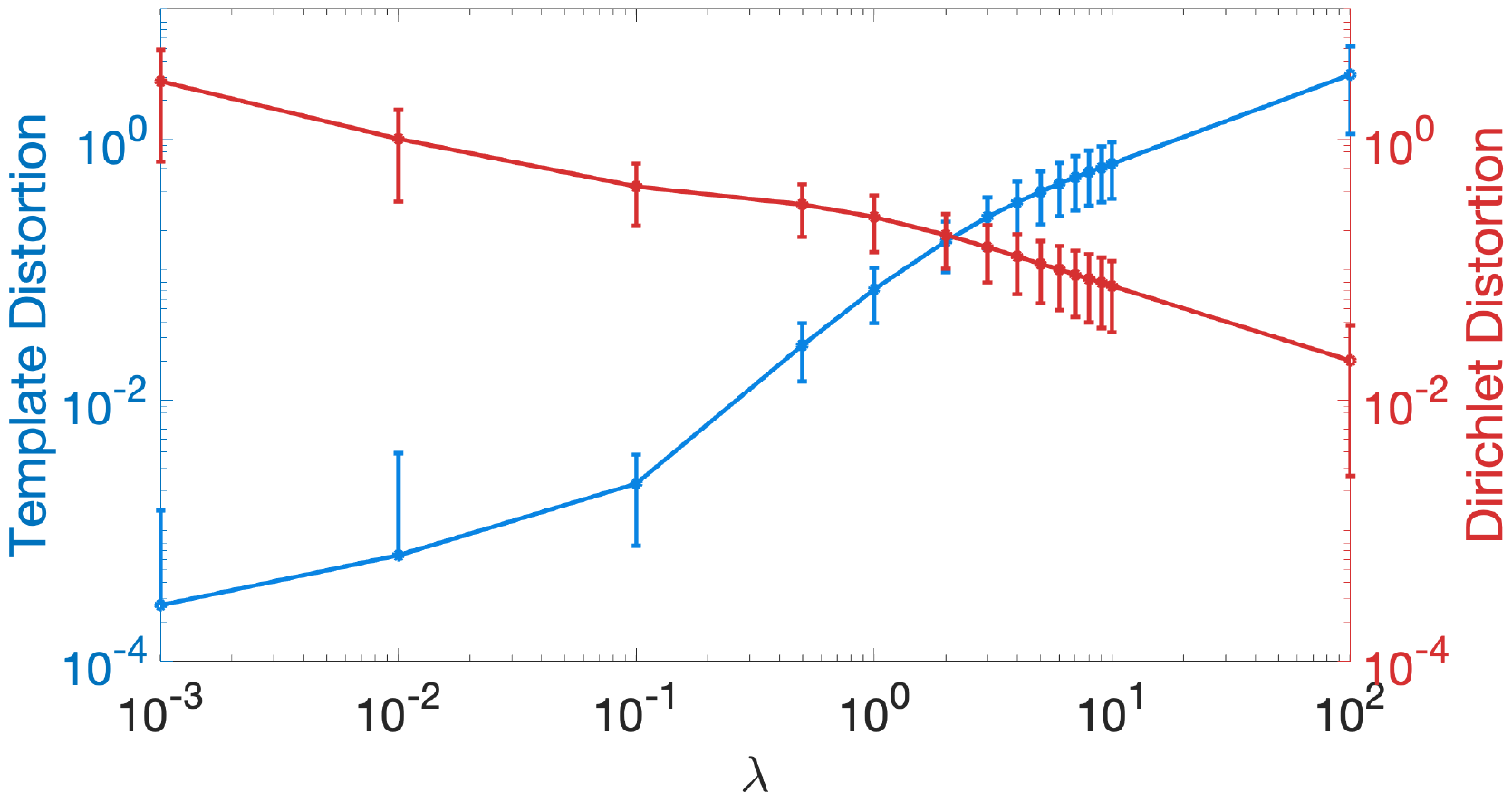}
\caption{The trade-off between template match and Dirichlet distortion for $\lambda \in [10^{-3},10^2]$, approaching optimality for $\lambda \in [0.1,5]$.}
\label{f:param-distortion}
\end{figure}

We employ grid search to set the values of the hyper-parameters. We set the shape distortion parameter $\lambda = 1$ as it was in the optimal trade-off range between the template fit and shape distortion, while producing a minimal final objective value. We observe that $\lambda \in [0.1, 5]$ produced similar results (Fig.~\ref{f:param-distortion}), thus the method is not very sensitive to the parameter $\lambda$. We initialize the template half-height $h$ and the ellipsoid radius $r_z$ as half of the placenta thickness, estimated from the \nth{95} percentile of the histogram of the distance transform values inside the segmentation boundary. We initialize the ellipsoid radius $r_y$ as half of the placenta width, estimated from the \nth{95} percentile of the histogram of shortest path geodesic distances along the placenta surface. We initialize the ellipsoid radius $r_x$ to maintain the placenta volume. We set the spectral clustering parameter $\gamma =20$ and use a boundary geodesic distance of five voxels (15 \textit{mm} in our application) as the half-width of the placenta margin. We set the line search parameters to $\beta = 0.9$, $\rho = 0.5$.  Our implementation is freely available at \url{https://github.com/mabulnaga/placenta-flattening}.

\section{Experimental Results}
\label{s:experiments}

In this section, we present our experimental results in data from a clinical research study to assess the quality of the transformation by visualizing the mapped BOLD MRI signals in the flattened space and quantify the resulting levels of distortion. We compare our approach to a 2D baseline parameterization method.

\subsection{Data}
\label{s:data}
We illustrate the utility of the proposed approach on a set of $111$ MRI scans acquired in 78 pregnant women. MRI BOLD scans were acquired on a 3T Skyra Siemens
scanner (single-shot GRE-EPI, 3 mm isotropic voxels, interleaved slice
acquisition, TR=5.8--8s, TE=32--36ms, FA= 90$^\circ$) using 18-channel
body and 12-channel spine receive arrays. There were a total of 18 twin and 60 singleton pregnancies, with gestational ages ranging from $26$ weeks, $6$ days to $37$ weeks, $6$ days. Scans of $33$ of the $60$ singleton pregnancy subjects were acquired in the supine and lateral positions, yielding $66$ of the total $111$ segmentations as the shape of the placenta changed between the two maternal positions. The twin pregnancies are monochornionic, meaning one placenta is shared by the twin fetuses.

The images were pre-processed to correct for signal nonuniformity and fetal motion during acquisition~\cite{turk2017spatiotemporal}. Following preprocessing, the placenta was manually
segmented by a trained expert. The resulting
segmentation label maps were used as input to the meshing software. The
resulting meshes had (mean $\pm$ std.) $6708\pm 1813$ tetrahedra and $2801\pm 731$ surface
triangles. 

\subsection{Method Variants and Baseline}
We evaluate our method on three proposed templates: \textit{ellipsoid}, \textit{parallel planes}, and \textit{single plane}. The single plane template is a variant of the parallel planes that allows the fetal vertices to relax while keeping the maternal vertices fixed to the bottom plane. This template is a relaxation of the parallel planes, and more closely mimics the \textit{ex vivo} examination process in which the maternal side is placed on an examination table. To implement this template, we first map the placenta to the parallel planes template and then run the optimization with a zero weight assigned to the fetal plane in the template term in of the optimization function. 

We compare our methods with a 2D parameterization approach. We emulate the prior placenta parameterization method of Miao \textit{et al.}~\cite{miao2017placenta} in which 2D curved surfaces spanning the placenta volume are independently flattened to a disk in $\mathbb{R}^2$ in a complex multi-step computational pipeline. These surfaces are constructed by cutting Euclidean distance level sets from the estimated medial axis of the placenta, with the surface boundaries defined to minimize local curvature change. We derive these 2D surfaces by intersecting the flattened volume with horizontal planes spaced one voxel apart. The intersection yields a set of parameterized surfaces. We map the surfaces back to the original placenta volume using the known barycentric coordinates, yielding a set of surfaces spanning the original placenta volume. We use harmonic parameterization~\cite{joshi2007harmonic} to map each surface to a disk. In order to align the disks, we chose the highest point in each surface to map to the north point of the disk. The radius of the disk can be selected arbitrarily, and we chose one that leads to zero-mean distributions of area and metric distortion. 

\subsection{Evaluation}
We evaluate the quality of the transformation by visualizing the mapped MRI signals and by measuring the quality of the template match and distortion. We visually assess the quality of our transformation by mapping the MRI intensity patterns to the template coordinate system. We examine the flattened mesh and anatomical landmarks of the mapped MRI signal in the flattened and original spaces. We compare our method with the 2D baseline by quantifying differences in areal distortion and visualizing the consistency in the mapped BOLD signal.   

We define the average template matching error as the square-root of the template matching term in Eq.~\eqref{eqn:objectivefun}, $(\sum_{m \in \partial Z}A_mT(x_m,\theta))^{\frac{1}{2}}$. We quantify shape distortion using the final Dirichlet distortion term in Eq.~\eqref{eqn:objectivefun}, and compute the fraction over the minimum of the symmetric Dirichlet energy for the identity transformation. We also examine the distribution of the log-determinant of the
Jacobian matrix $\log_2\det\left(J\left(X_k\right)\right)$ of tetrahedron $k$ to quantify local volumetric distortion~\cite{leow2007statistical}.  For comparisons with the 2D baseline, we compute area distortion using the ratio of triangle areas $\log_2(\hat{A}_m/A_m)$, where $A_m$ is the area of triangle $m$ in the original mesh and $\hat{A}_m$ is the area of the same triangle in the flattened mesh. We compute metric distortion using the ratio of edge length $\log_2\left(x_{ij}/z_{ij}\right)$, where $z_{ij}$ is the length of edge $(i,j)$ in the original mesh and $x_{ij}$ is the length of the same edge in the flattened mesh. Local metric distortion is used to measure how far the resulting map is from a local isometry and therefore how much it distorts the local signal patterns. Similarly, a volume (3D) or area (2D) preserving map is desirable to map the MRI signal intensities to the parameterized domain as described in Section~\ref{s:intensity-map}. 

\subsection{Results}
\label{s:results}

\subsubsection{Visual Results}
\begin{figure}[!t]
\centering
\includegraphics[width=0.4\textwidth]{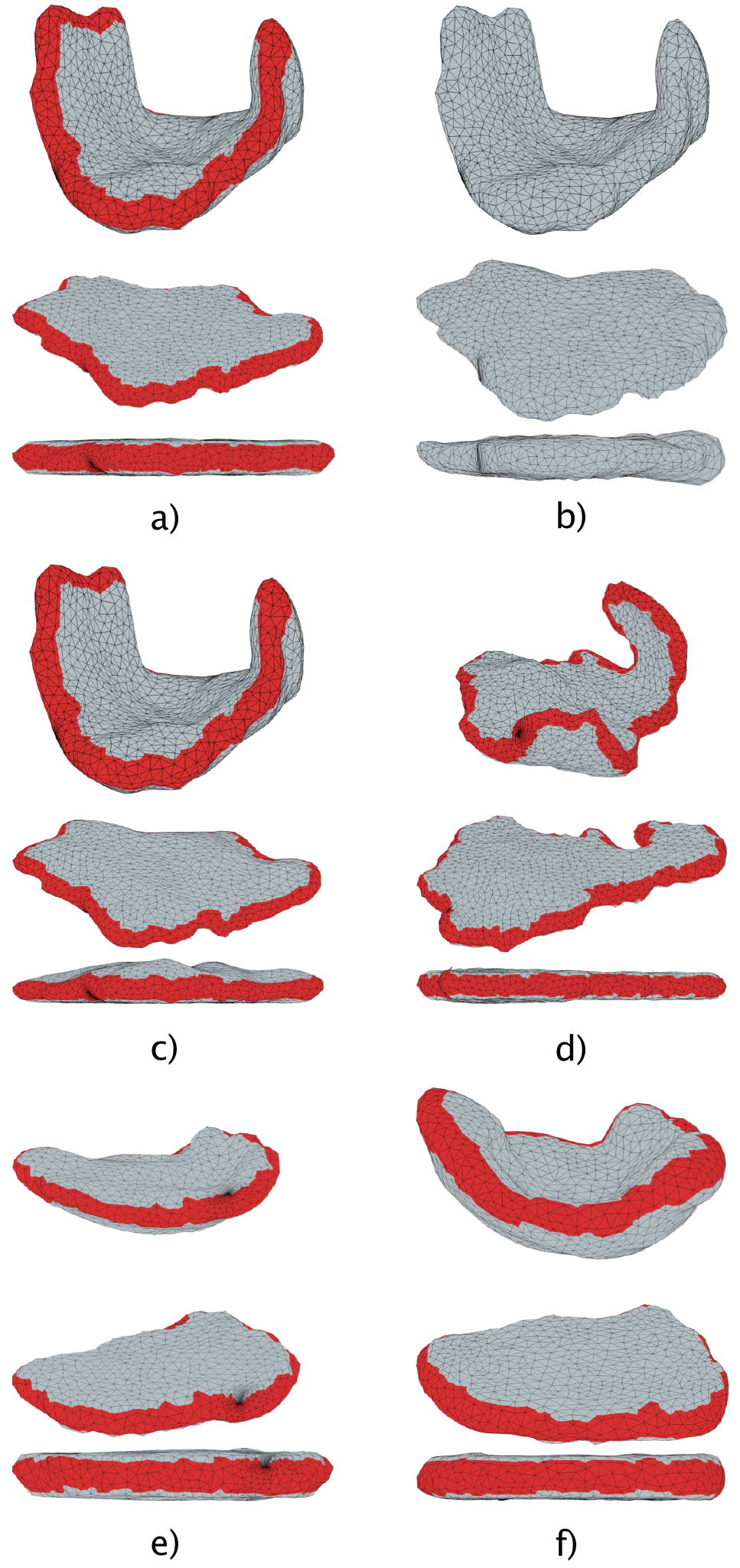}
\caption{Flattening results of tetrahedral meshes from three subjects. In each subfigure, the top row corresponds to the mesh in the original volume; the bottom row is the resulting flattened mesh. For each mesh using the parallel planes or single plane template, the region in red represents the placenta margin. The fetal side is facing upwards: (a) parallel planes; (b) ellipsoid; and (c) single plane flattening in a singleton subject; (d) parallel planes mapping in a twin subject. Sub-figures (e) and (f) demonstrate mapping the placenta as acquired in the supine and left lateral positions in a singleton subject using the parallel planes template. The ellipsoid template results in an unfolded shape though multiple curved areas in the fetal and maternal sides remain. The relaxation to the single plane template maintains the curvature of the fetal side (c). The same placenta acquired in two scan positions (e-f) appears visually more similar in the canonical template than in the original volume. }
\label{f:4subjects-mesh}
\end{figure}

Fig.~\ref{f:4subjects-mesh} illustrates the mapping results for four subjects. The algorithm enables robust mapping of the highly variable placental shapes encountered in the dataset. The ellipsoid template successfully unfolds the placenta, although curved regions remain throughout the volume. The parallel planes successfully unfolds the placentae and produces flat maternal and fetal sides. Mapping to the single plane template maintains the curvature in the fetal side of the organ while producing a flat maternal side. Mapping the placenta from the same subject acquired in two scanning positions is visually more similar in the canonical domain. The estimated placenta margin is connected and encapsulates the curved region that separates the fetal and maternal sides, highlighting the effectiveness of spectral clustering in parcellating the placenta boundary. The parallel planes template facilitates visualization of the maternal and fetal sides using the standard 2D cutting planes. Mapping using the parallel planes template results in meshes with a flat geometry which enables computation in the Euclidean domain. The remainder of our experiments focus on the parallel planes template. We perform quantitative comparisons of the three templates in Section~\ref{s:sub-quant-result}. We provide guidelines for template selection in Section~\ref{s:discussion}.

\begin{figure}[!t]
\centering
\includegraphics[width=0.49\textwidth]{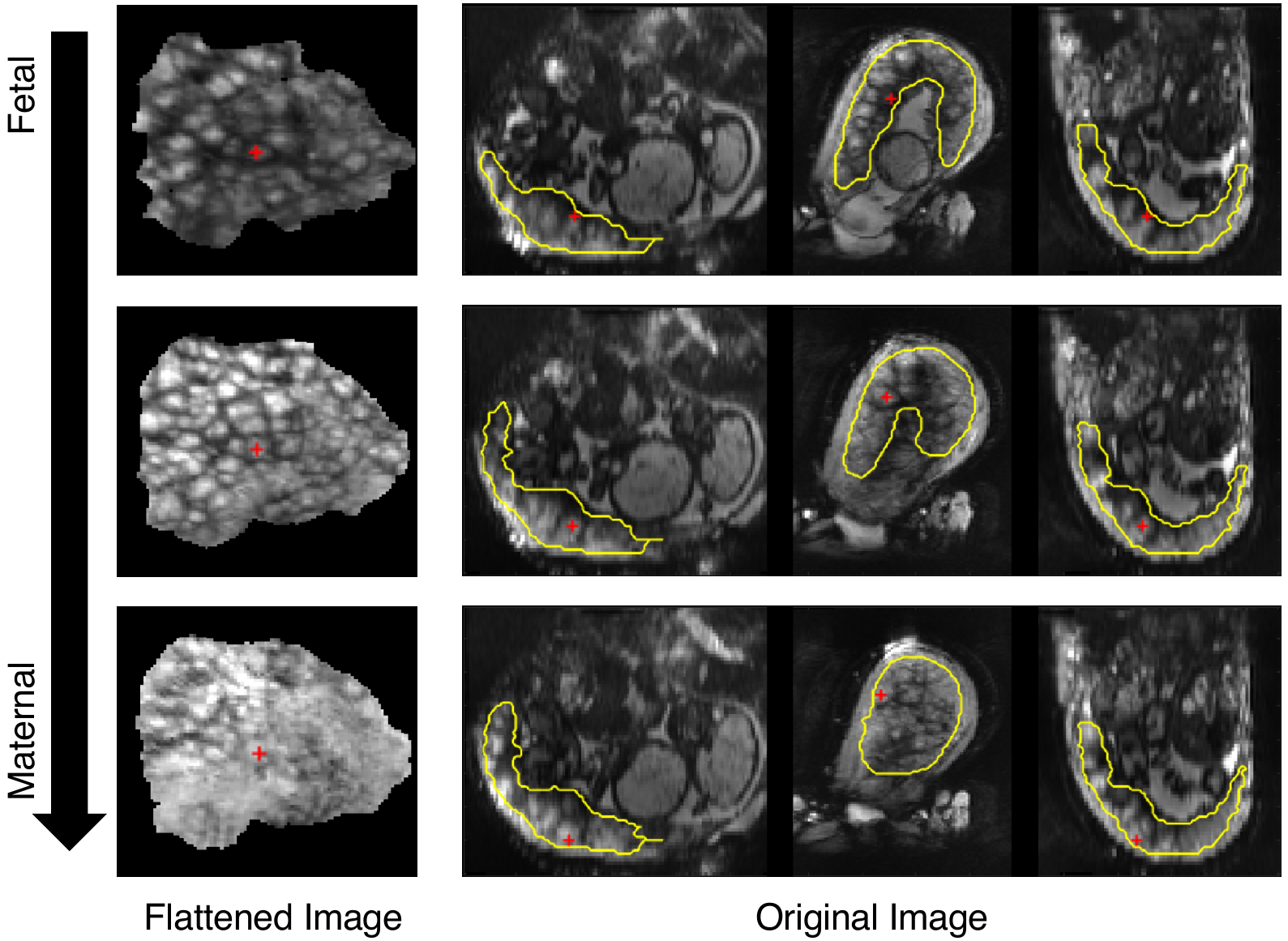}
\caption{ Visualizing anatomical landmarks in one twin pregnancy subject, moving depthwise from the fetal side (top) to the maternal side (bottom) of the placenta. Left column: flattened slices with mapped MRI signals. Right column: three orthogonal cross-sections of the original volume. Segmentation of the placenta is outlined in yellow, and the red marker outlines the approximate center of one cotyledon. The cotyledons, characterized by a honeycomb structure of small hyperintense regions, are immediately apparent in the flattened images. The spatial relationships between the marked cotyledon and others are lost in the original images.}
\label{f:cotyledon-flow}
\end{figure}

Fig.~\ref{f:cotyledon-flow} illustrates how the mapping enhances the visualization of landmarks. The mapped BOLD intensity patterns clearly visualize local anatomy as is apparent in the honeycomb structure of the cotyledons, which are the circular structures that support the exchange of oxygen and nutrients between the surrounding maternal blood and the fetal blood~\cite{benirschke1967pathology}. The cotyledons receive highly-oxygenated maternal blood through the spiral arteries which appear hyperintense in BOLD MRI, and receive less-oxygenated fetal blood through the chorionic villi, which appear darker. In the flattened space, we observe this effect as a darkening of intensity patterns closer to the fetal side. Our mapping brings forth the contextual information about the cotyledon distribution across the placenta, with consistent cotyledon structure maintained as we move from the fetal to the maternal side of the organ. The spatial relationship across cotyledons is difficult to observe in the original volume while the flattened view provides spatial information relevant for assessing placental function~\cite{luo2017vivo}. 

\begin{figure}[!t]
\centering
\includegraphics[width=0.49\textwidth]{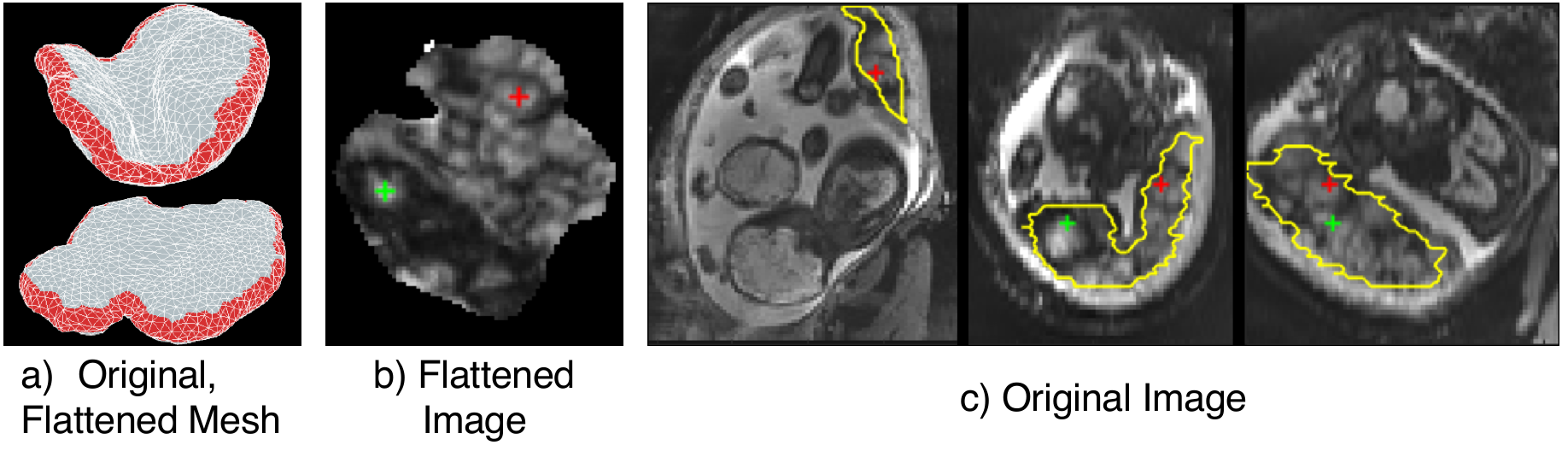}
\caption{ Visual evaluation of the spatial relationships between two cotyledons in a twin pregnancy subject. Left to right: (a) the original and flattened tetrahedral meshes; (b) one slice of the flattened volume; (c) three orthogonal cross-sections of the original image. The yellow outline marks the placenta boundary. The red and green ticks mark the approximate centers of two cotyledons. The views in the original image correspond to the center of the red cotyledon. We observe the spatial relationship between the two cotyledons in the flattened view clearly. Contextual information and spatial relationships are difficult to visualize in the original image due to the curvature of the uterine wall that determines the \textit{in vivo} shape of the placenta.}
\label{f:cotyledon}
\end{figure}

Fig.~\ref{f:cotyledon} demonstrates improved visualization of spatial relationships between landmarks in a twin subject. The flattened view provides contextual information surrounding the two marked cotyledons, e.g., the approximate distance between them and a clear delineation of two regions. The curved geometry of the \textit{in vivo} placenta makes it difficult to visualize the cotyledons together, losing the spatial relationships that are distinctly seen in the flattened view. As shown in Fig.~\ref{f:cotyledon-flow} and Fig.~\ref{f:cotyledon}, several cross-sections of the original image are required to identify the landmarks and their surrounding context.

\subsubsection{Quantitative Results}
\label{s:sub-quant-result}
Table~\ref{t:1-stats} reports the mean template matching error and Dirichlet distortion for each template. We observe sub-voxel accuracy of matching the template close to the minimum of the symmetric Dirichlet energy. We observe similar shape distortion between the ellipsoid and parallel planes templates, though the ellipsoid template has a larger template matching error with greater variance across subjects.

\begin{table}[!t]
\caption{Template mismatch and Dirichlet distortion per template.  }
\begin{center}
\begin{tabular}{||c || c| c | c||}
\hline
 \parbox[t]{20mm}{Cost Term\\ (mean$\pm$std. dev.)}& Ellipsoid & Parallel Planes & Single Plane \\
\hline
Dirichlet* (\%) & $3.4 \pm 2.6 $ & $4.2\pm 2.0$ & $2.0 \pm 1.0$ \\
\hline
Template (voxels) & $0.45 \pm 0.32$ & $0.26 \pm 0.05$ & $0.26 \pm 0.05$ \\
\hline
\end{tabular}
\label{t:1-stats}
\\
*The Dirichlet distortion is reported as a percentage over the minimum Dirichlet energy as given by the identity transformation.
\end{center}
\end{table}

\begin{figure}[!t]
\centering
\includegraphics[width=0.35\textwidth]{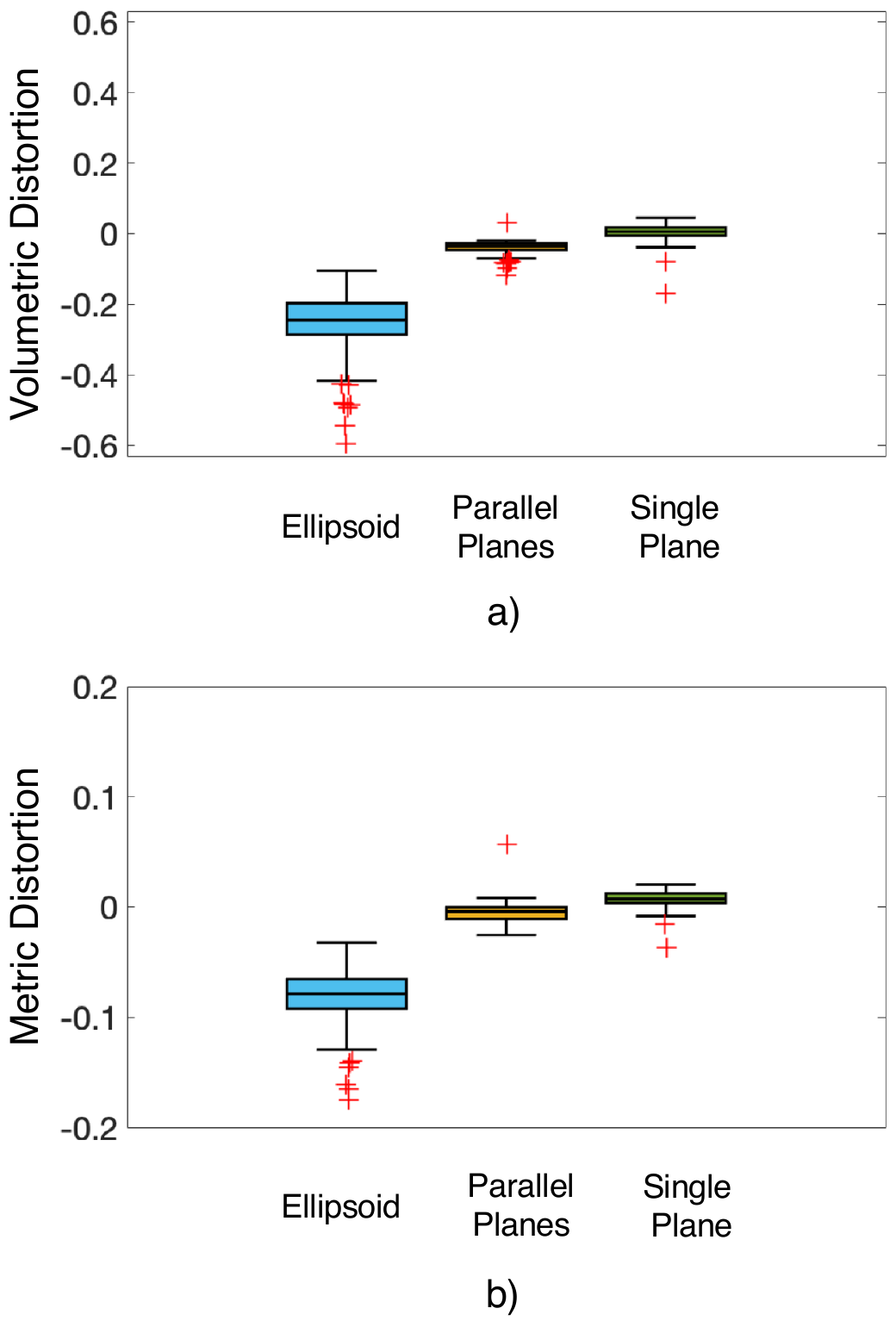}
\caption{ Box-and-whisker plots of (a) local volumetric $\left(\log_2\det\left(J\right)\right)$ and (b) metric $\left(\log_2\frac{x_{ij}}{z_{ij}}\right)$ distortions for the three templates. We report distributions of the mean distortion across the 111 cases. Black horizontal lines inside the box indicate the median, the boxes extend to the \nth{25} and \nth{75} percentiles, and the whiskers reach to the most extreme values not considered outliers (black bars). The outliers, shown as red crosses, are points farther than 1.5 times the interquartile range. Since both distortion measures are defined as $log_2$ of ratios, $1$ corresponds to doubling, and $-1$ corresponds to halving. The mean distortions are centered near zero, with narrow interquartile ranges and few outliers for the parallel planes and single plane templates. The ellipsoid template demonstrates shrinking in the mean distortion with a wider interquartile range.} 
\label{f:local-distortion}
\end{figure}

Fig.~\ref{f:local-distortion} reports the distributions of local volumetric and metric distortion measures for the three templates evaluated. We report distributions of the mean as the mean is sensitive to outliers and thus tight distributions of the mean values indicate well behaved measures across all subjects. Both the single plane and parallel planes templates yield narrow distributions of distortion centered at $0$ with few outliers. The ellipsoid template exhibits slight shrinking across the subjects and a wider interquartile range, indicating difficulty in some placentae conforming to an ellipsoid template. Given the large variability in placental shape, we observe that the parallel planes template more accurately models the desired flattened shape.

The optimization results in a template that more accurately models the placental thickness. We measure the absolute difference between the final and initial value of the template half-height $h, |\Delta h|= | h_{final} - h_{initial} | = 1.5\pm 1.0 \text{mm}$ for the parallel planes template. We observe $h_{final} =13.1\pm 2.6 \text{mm} $, which is in line with the reported average placenta thickness after delivery of $28\pm 5 \text{mm}$~\cite{barker2013maternal}.

We evaluate the robustness of our method to differences between singleton and twin pregnancy cases. For the parallel planes template, we compare distributions of local volumetric and metric distortion across the singleton and twin subjects using Welch's t-test to test if the distributions have the same means. The hypothesis is not rejected for the distributions of volumetric distortion ($p=0.16$) or metric distortion ($p=0.40$).

\begin{figure}[!h]
\centering
\includegraphics[width=0.48\textwidth]{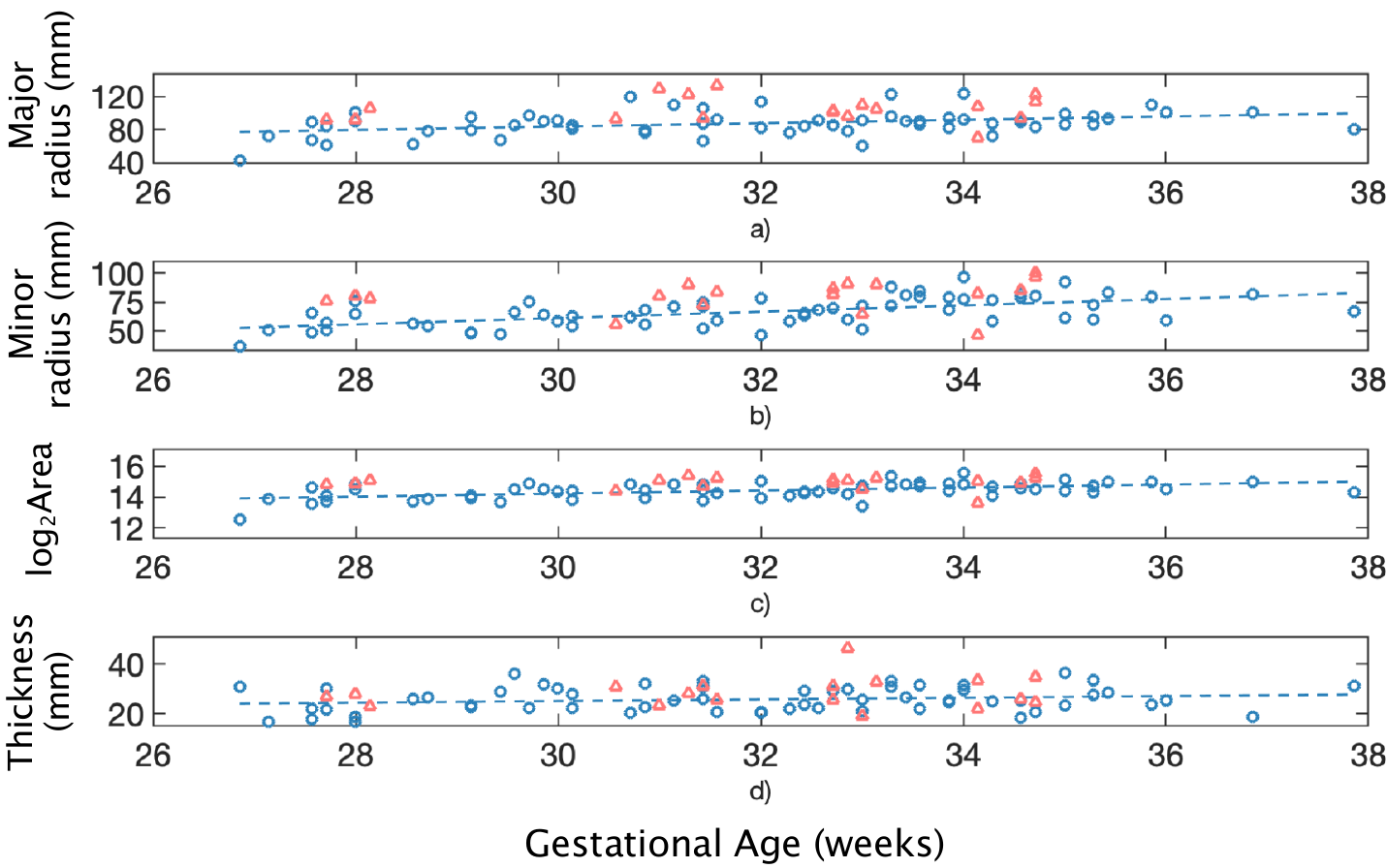}
\caption{Comparison of placental shape measurements in twin and singleton pregnancy scans: a)-c) major radius, minor radius, and the logarithm of the area of the central cross-section respectively; d) placenta thickness. Blue circles represent singleton scans and red triangles represent twin pregnancies. Dashed blue lines indicate linear regression fit for the singleton population. We conduct an ANCOVA analysis and find significant differences across the singleton and twin scans for the major radius, minor radius, and area, but not for the placenta thickness.}
\label{f:shape}
\end{figure}

To illustrate the benefits of the flattened representation, we examine placental shape differences between singleton ($n=60$) and twin ($n=18$) pregnancies. Specifically, we compute the major and minor radii and the area of the central cross-section, and the placental thickness in the flattened representation. The major and minor radii of the placenta are found by fitting the mesh vertices to an ellipse by minimizing the algebraic distance. We perform an ANCOVA analysis while accounting for gestational age as a covariate. Fig.~\ref{f:shape} shows scatter plots of these quantities as a function of gestational age. We found significant differences ($p<5\cdot\ 10^{-4}$) in the major and minor radii and the area, while differences in placental thickness were not significant. The same analysis using the fetal and maternal cross-sections yielded similar results. This study suggests that the placenta has larger extent but not substantially different thickness in twin pregnancies when compared to singleton pregnancies.

\begin{figure}[!h]
\centering
\includegraphics[width=0.35\textwidth]{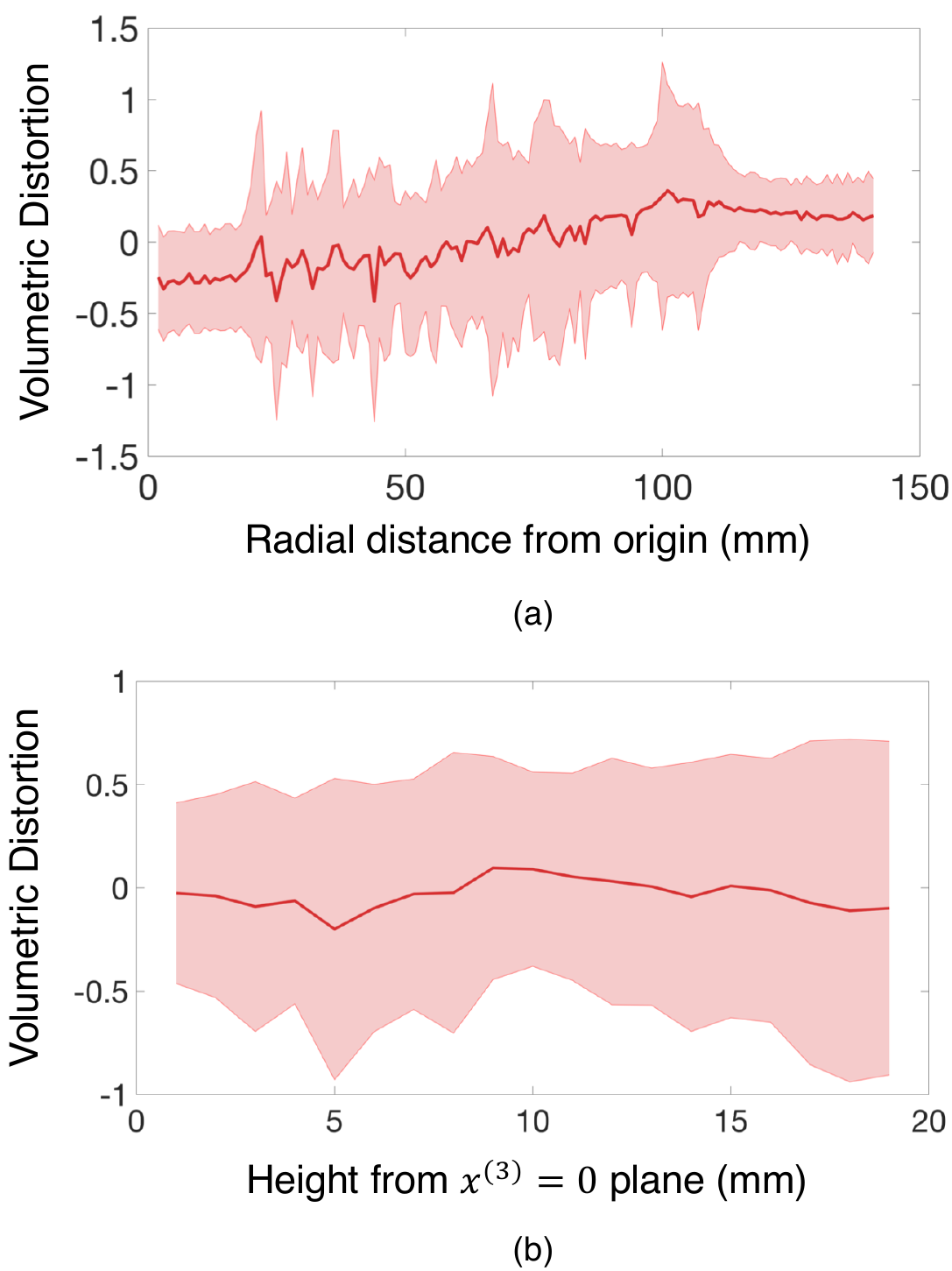}
\caption{ Local volumetric distortion as a function of vertex distance: mean volumetric distortion as a function of (a) vertex's planar distance from the origin $((x^{(1)})^2 + (x^{(2)})^2)^{\frac{1}{2}}$ and (b) absolute vertex height from the $x^{(3)}=0$ plane. Solid line indicates mean distortion and shaded areas extend to $\pm 1$ standard deviation across all vertices in the 111 cases. There are small variations in distortion as a function of distance, indicating distortion is evenly distributed throughout the volumes. In (a), we observe shrinking near the origin, which is expected, given the positively curved placental shape.}
\label{f:distance-distortion}
\end{figure}

We also evaluate the spatial distributions of distortion. Fig.~\ref{f:distance-distortion} reports volumetric distortion as a function of the vertex distance from the origin (x-y plane) and vertex height from the $x^{(3)}=0$ plane, indicating distortion is evenly distributed throughout the placenta volumes.

\subsubsection{Comparison to 2D Baseline}

\begin{figure*}[h!]
\centering
\includegraphics[width=0.65\textwidth]{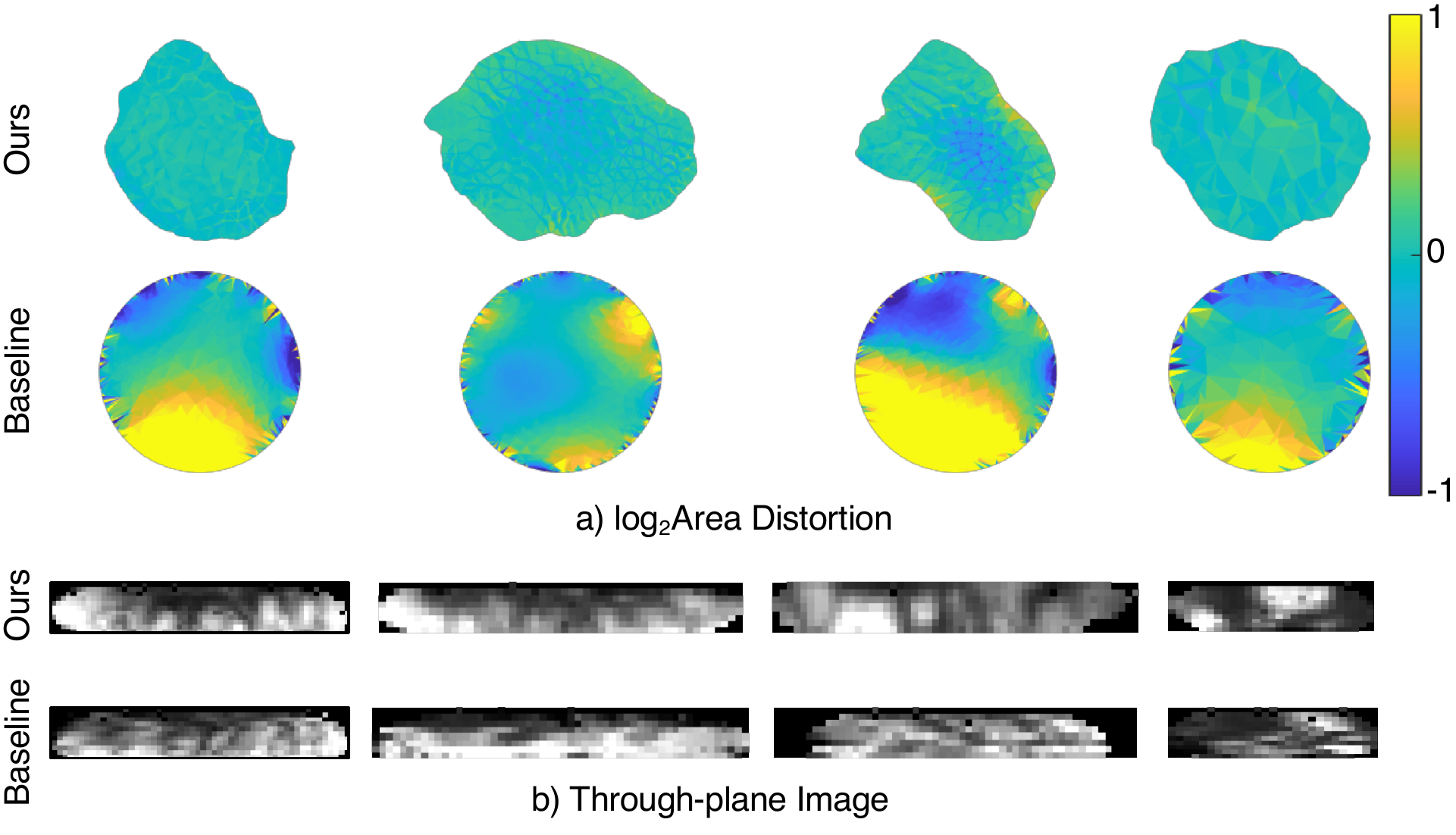}
\caption{ Areal distortion and image mapping for our method and the baseline 2D method, demonstrated for four subjects: (a) distributions of areal distortion on a central plane; (b) through-plane view of the mapped MRI values in the parameterized space by our method (top) and the baseline 2D method (bottom). Our method results in more homogeneous distributions of distortion. The baseline severely distorts the placenta at the boundary. The through-image view of the mapped intensities demonstrates intensity artifacts due to independent parameterization of different planes.  }
\label{f:2d-intensity}
\end{figure*}

Fig.~\ref{f:2d-intensity} compares the distributions of areal distortion and the mapped intensity patterns for four subjects. The baseline method of parameterizing to a disk produces high distortion near the boundary, and introduces misalignment of the flattened surfaces, as is visible by the through-plane images. The baseline 2D mapping parameterizes the surfaces independently of each other, lacks coupling across surfaces, and produces strong intensity artifacts. These results support the need for a volumetric parameterization method. 

\begin{figure}[h!]
\centering
\includegraphics[width=0.3\textwidth]{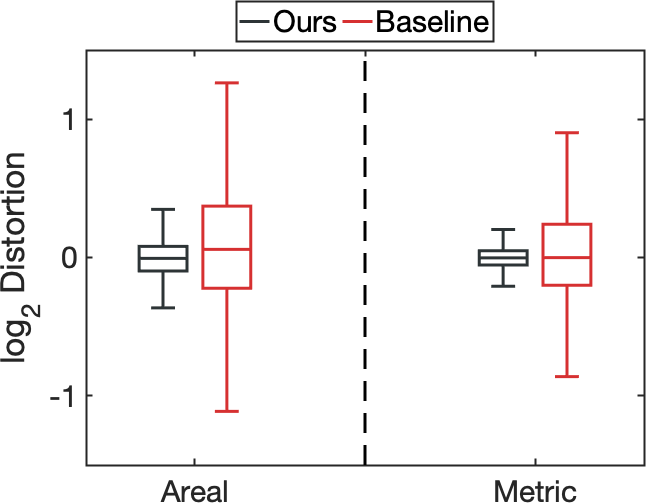}
\caption{ Areal and metric distortion, for our method and the 2D baseline parameterization method. The distributions are across the triangles (areal) and edges (metric) of all parameterized surfaces. }
\label{f:2d-whisker}
\end{figure}

Fig.~\ref{f:2d-whisker} compares our method's distributions of areal and metric distortion with the baseline 2D approach. We achieve significantly lower distortion across all cases. We conduct a two-sample Kolmogorov-Smirnov test to assess the signficance. Our method produces significantly lower areal and metric distortion per placenta ($p<10^{-24}$).

\section{Discussion}
\label{s:discussion}
Our volumetric parameterization approach successfully maps the placenta to a canonical template that closely resembles the \textit{ex vivo} shape, thereby enhancing visualization of anatomy and function. Experiments comparing cotyledon distributions in the flattened and original views demonstrate that spatial relations of anatomical landmarks are more clearly seen, and fewer slices are required to span the placenta volume. Both local and global distortions are successfully controlled while ensuring good template fit. Our algorithm is robust to the greatly varying placental shapes encountered in the study, successfully handling both twin and singleton cases. Our volumetric approach results in significantly less distortion and ensures consistent parameterization throughout the volume to a greater extent than that of a baseline 2D parameterization approach. The significant differences in distortion are due to the baseline approach fixing the placental boundary to a disk. The natural placental boundary is often not disk-like, as can be seen in Fig.~\ref{f:4subjects-mesh}, resulting in large distortion. Our approach parameterizes the placenta to a free boundary, thereby maintaining the natural shape, and is therefore better suited for placental parameterization.

We evaluate three templates, the ellipsoid, parallel planes, and single plane. The proposed parallel planes template results in a parameterization with less distortion compared to the ellipsoid. The single plane, which is a modification of the parallel planes, results in slightly lower distortion. The single plane template matches closely the \textit{ex vivo} examination process as the curvature in the fetal side is maintained. The parallel planes template is more natural for visualization using standard 2D cutting planes. The flexibility in the proposed approach enables the user to select a desired template. We recommend the parallel planes template for most uses. In addition to its advantages for visualization, the parallel planes template enables computation in the Euclidean domain by producing flattened geometry within the placenta. The single plane template minimizes local distortion and is recommended for regional analysis within the volume of the placenta.

Our results demonstrate improved visualization of the BOLD MRI signals. Since our model is geometry-based, it can also be used with other modalities, provided the placenta segmentation is given. Additional modalities, such as diffusion MRI~\cite{slator2019combined} and T2-weighted MRI~\cite{torrents2019fully} are used to study placental function and anatomy. This provides an opportunity to map and register several signals to the common template for visualization and analysis. 

An alternative and natural formulation of this work would be to use a physical modeling approach, where the parameterization is modeled as a physical deformation under gravity. Unfortunately, to the best of our knowledge, there is minimal published work on the tissue properties of the placenta. The symmetric Dirichlet energy is an elastic energy and is an appropriate proxy for a physical deformation. The formulation of our template as two parallel planes was inspired by a physical system, where the bottom plane (maternal side) is the post-delivery examination table, and the top plane forces the placenta to flatten. By optimizing for the template height, the optimization results in a template that minimizes compressive and expansive forces on the placenta.

\subsection{Limitations and Future Work}
A limitation of our template formulation is the invariance to rotation. Rotations about the $x^{(3)}-$axis do not affect the value of the objective function, which presents difficulty for aligning placentae, or the placenta of the same subject over time. Additional registration may be required for proper alignment. Furthermore, our template does not provide detailed correspondence. For example, although we map the maternal side to the bottom plane, we are unable to distinguish the left or right side of the placenta. This is a challenging problem that will motivate the future methodological development.

Going forward, we plan to improve the estimation of the placenta margin. Currently, we overestimate the margin to prevent distortion caused by incorrectly assigning vertices to the maternal or fetal side. An improvement is to refine the margin size throughout the optimization, converging on the margin that estimates the placental thickness.

We plan to improve the template using anatomical landmarks such as the umbilical cord insertion. The cord cannot be assumed to be inserted in the center of the placenta, especially in twin cases, and must be identified in the image. We are currently collecting anatomical images to identify landmarks  as they are not easily seen in the lower-resolution BOLD (functional) images. The use of anatomical landmarks is necessary for the development of a common coordinate system of the placenta. A potential approach to refining correspondences between the placenta and the template  would be to co-register the anatomic and functional images prior to flattening, then compute spatial statistics using mapped landmarks in the flattened space to enable comparative studies.

We envision several potential clinical research studies enabled by this work. The flattened view can be used to visualize regional MRI signal differences, for example within cotyledons, the cord insertion, and the placental regions responsible for each twin pair. Visualization of regional differences will support quantification of signal differences, for example across discordant twins with IUGR. Further, the canonical domain enables studies comparing shape differences across populations as in Fig.~\ref{f:shape}. In Fig.~\ref{f:cotyledon-flow} and Fig.~\ref{f:cotyledon}, we demonstrate visualization of the spatial relationships across cotyledons. Assessing the signal variations across cotyledons is a promising direction to study the possible conditions during pregnancy and after birth. For example, the number of cotyledons at birth has been shown to be a predictor of blood pressure in childhood~\cite{barker2013maternal}.

In future work, we plan to quantify the similarity between the proposed parameterization and the \textit{ex vivo} placental shape using, for example, an \textit{ex vivo} system for placental perfusion and MR imaging~\cite{stout2020placental}. By imaging a pregnant mother near gestation and comparing the flattened placenta with the \textit{ex vivo} images, we can quantify the similarities between the two. A strong correlation between the \textit{in vivo} and \textit{ex vivo} shapes will enable pathology sampling of potential biomarkers identified during pregnancy. Furthermore, characteristics of placental shape have been shown to correlate with pathology including preeclampsia~\cite{kajantie2009preeclampsia}, and conditions that affect the fetus later in life such as hypertension~\cite{barker2010surface} and coronary heart disease~\cite{eriksson2011mother}.

\section{Conclusion}
\label{s:conclusion}
We developed a volumetric mesh-based mapping of the placenta to a flattened template represented by two parallel planes, resembling its \textit{ex vivo} flattened shape. Our method successfully parameterizes the placenta with minimal distortion and guarantees local injectivity. In a clinical study of placental BOLD MRI images, we demonstrated that the mapping improves visualizing local anatomy and function, and its potential for becoming a standardized method for representing and visualizing placental images.

\bibliographystyle{ieeetr}
\bibliography{main}
\begin{appendices}
\section{}
\label{appendix:a}

\subsection{Jacobian Formulation}
\label{appendix:jacobian}
The Jacobian defined in Eq.~\eqref{eqn:jacobian} defines the transformation of tetrahedron $k$ using the three basis vectors $X_kB$ defining the tetrahedron. The matrix $B\in\mathbb{R}^{3\times 4}$
 \[B =\begin{bmatrix}
    \ -1     & -1 & -1\\
    1     & 0 & 0  \\
    0      & 1 & 0 \\
    0      & 0 & 1
\end{bmatrix}\]
is a linear operator that extracts the basis vectors. Matrix $X_k$ represents the vertices of tetrahedron $k$, $\{x_i,x_{i+1},x_{i+2},x_{i+3}\}$ as column vectors. The basis used to compute the Jacobian is then

\[X_kB = 
 \begin{bmatrix}
    \vert  & \vert & \vert  \\
    x_{i+1}-x_{i}  & x_{i+2}-x_i & x_{i+3}-x_i \\
    \vert  & \vert & \vert 
\end{bmatrix}. \]

\subsection{Gradient of the Objective Function}
\label{appendix:gradient}
The gradient of the objective function in Eq.~\eqref{eqn:objectivefun} per vertex is computed as the sum of the gradient of the template and the distortion terms. The gradient of the parallel planes template term with respect to the boundary vertex $m$ is linear, 
\begin{equation*}
\frac{\partial T(x_m,h)}{\partial x_m}  =
\begin{cases}
2(x_m^{(3)} - h) & \text{if } m \in \mathcal{F}(\partial Z), \\
2(x_m^{(3)} + h) & \text{if } m \in \mathcal{M}(\partial Z), \\
0 & \text{otherwise}.
\end{cases}
\end{equation*}

The gradient of the ellipsoid template term with respect to boundary vertex $m$ is

\begin{equation*}
\frac{\partial T(x_m,R)}{\partial x_m}  = 4A_m(x_m^T  R x_m - 1) R x_m .
\end{equation*}

The gradient of the distortion term for tetrahedron $k$ is:
\begin{align}
\frac{d\, \|J(X_k) \|_F^2}{d\, X_k} & = 
X_k \left[ 2 B \left(Z_{k} B\right)^{-1} \left(Z_{k} B \right)^{-T} B^T\right], \nonumber \\
\begin{split}
\frac{d\, \|J(X_k) ^{-1}\|_F^2}{d\, X_k} &= -\left [
 2 X_k B \left(B^T X_k^T X_k B \right)^{-1} B^T Z_k^T Z_k B\right ] \times \\
 &\quad \quad \quad \quad \left [ \left(B^T X_k^T X_k B \right)^{-1} B^T \right]. \nonumber
 \end{split}
\end{align}

We also optimize over the template height. For the two parallel planes template, the corresponding derivative is

\begin{equation*}
\frac{\partial \phi(X,h)}{\partial h} = -2\left [\sum_{j \in \mathcal{F}(\partial Z)}A_j \left(x_j^{(3)}-h\right)  - \sum_{j \in \mathcal{M}(\partial Z)}A_j \left(x_j^{(3)}+h \right)  \right]. \\
\end{equation*}

For the ellipsoid template, the corresponding derivative with respect to the parameter $r_x$ is

\begin{equation*}
\frac{\partial \phi(X,R)}{\partial r_x} = -\frac{4}{r_x^3}\sum_{m \in \partial Z}A_m \left(x_m^TRx_m-1\right). \\
\end{equation*}
An equivalent form is found for $r_y$ and $r_z$.
\end{appendices}
\end{document}